\documentclass[nohyperref]{article}

\usepackage{microtype}
\usepackage{graphicx}
\usepackage{subfigure}
\usepackage{booktabs} 

\usepackage{hyperref}

\usepackage[accepted]{icml2022_TAGML}
\usepackage{amsmath}
\usepackage{amssymb}
\usepackage{mathtools}
\usepackage{amsthm}

\usepackage[capitalize,noabbrev]{cleveref}

\theoremstyle{plain}

\theoremstyle{definition}

\theoremstyle{remark}

\usepackage[textsize=tiny]{todonotes}

\icmltitlerunning{Rethinking Persistent Homology for Visual Recognition}

\begin{document}

\twocolumn[
\icmltitle{Rethinking Persistent Homology for Visual Recognition}



\icmlsetsymbol{equal}{*}

\begin{icmlauthorlist}
\icmlauthor{Ekaterina Khramtsova}{uq}
\icmlauthor{Guido Zuccon}{uq}
\icmlauthor{Xi Wang}{neusoft}
\icmlauthor{Mahsa Baktashmotlagh}{uq}
\end{icmlauthorlist}

\icmlaffiliation{uq}{University of Queensland , Brisbane, Australia}
\icmlaffiliation{neusoft}{ Neusoft, China}

\icmlcorrespondingauthor{Ekaterina Khramtsova}{ekaterina.khramtsova@uq.edu.au}

\icmlkeywords{Machine Learning, ICML}

\vskip 0.3in
]



\printAffiliationsAndNotice{}  

\begin{abstract}
Persistent topological properties of an image serve as an additional descriptor providing an insight that might not be discovered by traditional neural networks. The existing research in this area focuses primarily on efficiently integrating topological properties of the data in the learning process in order to enhance the performance. However, there is no existing study to demonstrate all possible scenarios where introducing topological properties can boost or harm the performance. 
    This paper 
    performs a detailed analysis of the effectiveness of topological properties for image classification in various training scenarios, defined by: the number of training samples, the complexity of the training data and the complexity of the backbone network.
    We identify the scenarios that benefit the most from topological features, e.g., training simple networks on small datasets. Additionally, we discuss the problem of topological consistency of the datasets which is one of the major bottlenecks for using topological features for classification. We further demonstrate how the topological inconsistency can harm the performance for certain scenarios.
    
\end{abstract}

\section{Introduction}

\label{sec:intro}

Topological Data Analysis (TDA) is a mathematical framework that aims to extract robust geometrical and topological information from high-dimensional data. Recently TDA has experienced an increasing interest from the machine learning community and has proven valuable in a broad range of applications, including shape recognition \cite{hofer_deep_2018},  graph \cite{carriere_perslay_2020} and image classification \cite{dey_improved_2017}, adversarial attacks and model interpretability \cite{moor_topological_2020}.

One of the most common methods of calculating robust and stable topological features is persistent homology \cite{edelsbrunner_2010}, \cite{zomorodian_2005}. In a nutshell, persistent homology monitors the evolution of the topology of the space throughout a filtration. The output of persistent homology represents a multiset, that summarizes the life and death of topological invariants of different dimensions (i.e., 
connected components, holes).


A naive way of integrating topological properties into the machine learning pipeline is to apply persistent homology on each input sample and approximate the resulting multiset to form a fixed-sized vector ~\cite{qaiser_fast_2018, chung_topological_2018}.
The drawback of this approach is that the vectorization of the multi-set is non-differentiable, which makes the topological representations task-agnostic. To overcome this limitation, efficient ways of utilizing topology are proposed which work by either imposing certain topological priors through the loss function~\cite{clough_topological_2020,byrne_persistent_2020}, or by integrating topology as a layer in neural network~\cite{hofer_deep_2018,carriere_perslay_2020}: the methods of \cite{clough_topological_2020,byrne_persistent_2020} enforce segmentation masks to have a certain number of connected components and holes~\cite{clough_topological_2020,byrne_persistent_2020}; the method of \cite{bruel-gabrielsson_topology_2019} imposes generated images to have a predefined topological structure; and proposed recently, the state-of-the-art topological network layer~\cite{hofer_deep_2018,carriere_perslay_2020, kim_efficient_2020} integrates the topology as a fully differentiable layer which can be placed in any part of the main network. 

The methods that rely on imposing the topological priors are limited in the sense that the ground truth for the topology of the dataset is not always available in practice, except for some segmentation masks \cite{clough_topological_2020}, or for  simplistic datasets such as MNIST. Moreover, all 
the existing methods rely on the assumption that the datasets are topologically consistent. In the context of image classification, this can be translated to a topological inter-class heterogeneity and intra-class homogeneity. In other words, the images that belong to the same class are assumed to be topologically similar, while being topologically distinct from the other classes. 
Topological consistency depends on the correct choice of the complex and a filtration, which requires a solid understanding of the persistent topological structures of the samples. 

In light of the above discussion, in this paper, we analyse the implications of having a topological invariance and discuss how to overcome this issue. 
We extensively study the effect of utilizing topological features for the task of image classification with respect to the dataset size and complexity, and backbone network complexity.

In doing so, we make the following contributions:

\begin{itemize}
    \item 
    We show that the influence of incorporating the topological layer in simple neural network architectures is bigger compared to the deeper networks, resulting in a noticeable performance improvement for the task of image classification.
    
    \item We analyse the importance of persistent topological features with respect to the dataset size. Demonstrated by our empirical results, when only a limited number of training samples is available, persistent topological features can significantly improve the overall performance of the classifier. As more data becomes available, the influence of the topological features on the main classifier decreases. 
    
    \item We show that the classification of the images with more topologically diverse structure (e.g. medical images) can substantially benefit from incorporating topological features, when only a few training samples are available.
    This finding is particularly important for classification of rare diseases that have limited number of samples. 
    In contrast, adding the layer to the main network might become harmful once there is enough data available. 
   
\end{itemize}

\vspace{1ex}
\section{Related work}
\label{sec:related}
 
\begin{figure*}
\rule{0pt}{2in}
   \includegraphics[width=0.95\textwidth]{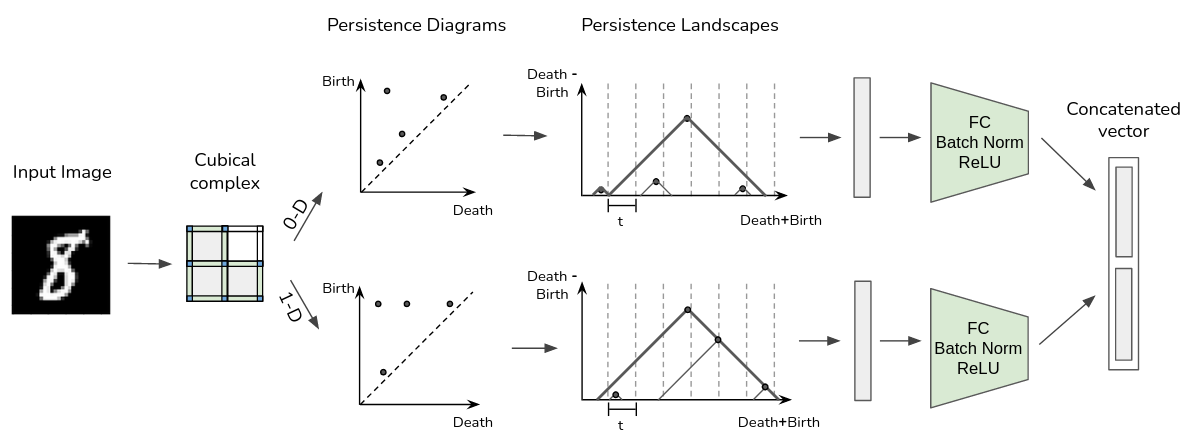}
     \caption{
     The architecture of the Landscape layer. A filtered cubical complex is constructed from the input image. Persistent homology is further applied to extract topological features, resulting in 0-D and 1-D persistence diagrams. Then, persistence landscapes are calculated to vectorize the multiset of points from persistence diagrams. The resulting  fixed-sized 
     vectors are separately fed into two Linear layers, each followed by Batch Normalisation and ReLu activation function. Finally, the output of the layers is concatenated to form the final vector.     
   }
   \label{fig:landscape_layer} 
   \vspace{-1ex}
\end{figure*} 
\noindent \textbf {Topological statistics:} The first efforts to introduce topological properties of the input samples to neural networks were in calculating various statistics on the persistence diagram and use them to train a stand-alone machine learning model.
For example, Chung et al. 
\yrcite{chung_topological_2018} train a SVM on persistence curves and persistence statistics for a skin disease analysis.
Quaiser et al. \yrcite{qaiser_fast_2018} train a Regression Forest on Betti curves. They further design an ensemble strategy to combine the predictions from the regression forest and a convolutional neural network and apply on patch classification and tumour segmentation. 


The main drawback of topological statistics is that the transformations for their calculation are non-differentiable, and therefore the network cannot be trained end to end. 


\vspace{1ex}
\noindent \textbf {Loss function:} A more efficient way to benefit from topological information is to enforce some ground truth knowledge on the receptive space. For example, in image segmentation with one object Clough at al.
\yrcite{clough_topological_2020} correct the masks generated by the autoencoder by combining MSE loss with topological loss, that enforce the mask to have one connected component and one hole. Byrne et al. \yrcite{byrne_persistent_2020} similarly use the topological loss to correct segmentation masks for multi-class cardiac MRI. 

A more generalised approach for segmentation correction was proposed by Hu et al.
\yrcite{hu_topology-preserving}, where the authors reduce the difference between the topology of the generated mask and those of the ground truth mask by minimizing a modified Wasserstein distance between their persistence diagrams.

Segmentation correction is not the only task, that can benefit from topological loss. For example, Moor et al.\yrcite{moor_topological_2020} 
propose to minimize the difference between the topology of the input and latent spaces of the autoencoders. They argue that preserving the topology of the latent space helps improving interpretability of the autoencoder.

Finally, topological loss can be used in the context of noise reduction and data regularisation. Bruel-Gabrielsson et al. 
\yrcite{bruel-gabrielsson_topology_2019} demonstrate that imposing some topological priors helps to improve the reconstruction loss. 

The biggest drawback of the topological loss function is that it requires a ground-truth knowledge of the topology of the receptive space, that is realistically only available for very simplistic datasets \cite{bruel-gabrielsson_topology_2019} or for segmentation masks.  

\vspace{1ex}
\noindent \textbf{Topological layer:} The first topological layer is described in detail by Hofer et al.
\yrcite{hofer_learning_rep}. The authors focus on a variant of Gaussian transformation on points from persistence diagrams and compare different projection types. 
The layer is used for graph classification and 2D and 3D shape recognition. Differently from other works, that construct Cubical complex for visual input, the authors define a heat-kernel signature of the object’s surface mesh and apply a height function to create a filtration. 

Zhao et. al. \yrcite{10.5555/3454287.3455171}  build a weighted kernel, constructed on persistence images representation and demonstrate its efficiency on a graph classification task.
Kim et al.\yrcite{kim_efficient_2020} propose to use topological layer to improve the classification of noisy images. They construct a Cubical complex from input images and use persistence landscapes as a vectorization method. The biggest contribution of their work is the use of DTM (Distance to measure) function for creating a filtration.

Finally, Carri\`ere et al.
\yrcite{carriere_perslay_2020} extend previous works by generalizing various vectorization methods for persistence diagrams, that they call PersLay. The authors construct a heat Kernel Signatures (HKS) of a graph and use Perslay for graph classification.

\section{Background}

In this section, we will go through some basic concepts and definitions of topological data analysis (TDA).

\noindent \textbf{Cubical Complex:} 
 Images consist of pixels, that can also be seen as 2D cubes and can thus be represented with a Cubical complex. It can be defined as follows: 
    
\textit{An elementary interval} is an interval of the form $I = [k,k+1]$ or the degenerate interval $I=[k,k]$ for some $k \in \mathbb{Z}$.
    \textit{An elementary cube} is a Cartesian product of elementary intervals: $Q= I_1 \times ... \times I_d $.
    A set $K$ is a \textit{cubical complex} if it can be written as a finite union of elementary cubes. Naturally, for 2D images each pixel $X[i,j]$ is filled in with the elementary cube $Q_{i,j}=[i,i+1]\times[j, j+1]$, $Q_{i,j} \in K_2$. 
    
    
\noindent \textbf{Persistent Homology:}    Let $K$ be a simplicial complex. A sequence of simplicial complexes, such that $\emptyset = K_0 \subset ... \subset K_n = K$ is called a \textit{filtration} of $K$. For a cubical complex the filtration is defined by increasing the threshold $\alpha \in [$min$(X), $max$(X)] $ and constructing cubical complexes for each value of $\alpha$ as follows:
    $K_\alpha = \{Q_{i,j} : X[i,j]> (1-\alpha)\}$
    
Persistent homology is a method that keeps track of topological structures that persist throughout a filtration. 
Our input is 2D, therefore we restrict the analysis to 0D and 1D homological features, aka connected components and holes. 
    
\noindent \textbf{Persistence Diagrams:} The resulting homology of the filtration is summarised in a persistence diagram $\mathcal{D} = \{\mathcal{D}_0, \mathcal{D}_1\}$, where each point $(b_i, d_i)$ indicates that a topological feature appeared when the threshold $\alpha = 1- b_i$ and disappeared when $\alpha = 1-d_i$. 
    In other words, the feature was \textit{born} at scale $b_i$ and \textit{died} at scale $d_i$. 
    

    The points from persistence diagrams represent a multi-set and therefore are not directly suitable as an input to a Neural Network. For this reason they need to be vectorized first.
    
    
\noindent \textbf{Diagram Approximation:} There exist various diagram approximation methods ~\cite{adamce_persistence_image, chazal_persistence_silhouettes}). One approach is to discretize the domain into fixed-sized bins and calculate the so called persistence landscape \cite{bubenic_landscapes}.
     
     First, we define the binning interval T$=[t_0,...,t_q] \in \mathbb{R}$ of equidistant points: $t_{n+1}-t_n=t$ $\forall n \in \mathbb{N}$, where $t_0=$ min$(\alpha), t_q=$ max$(\alpha)$. Then each point $p=(b_i, d_i) \in \mathbb{D}$ of the persistence diagram is mapped via the triangle point transformation as follows: $p \rightarrow [\Lambda_p(t_0), ..., \Lambda_p(t_q)]$; $\Lambda_p: t_i \rightarrow $max$(0, y-|t_i - x|)$. Once all the points are mapped to $\mathbb{R}^q$, the values for each interval are  sorted and only the $k$ largest   values for each binning point are kept. It is denoted as top-$k$ persistence landscape.

\section{Methodology}

\begin{figure*}
   \vspace{-3ex}
\begin{tabular}{cc}
\includegraphics[width=0.42\linewidth]{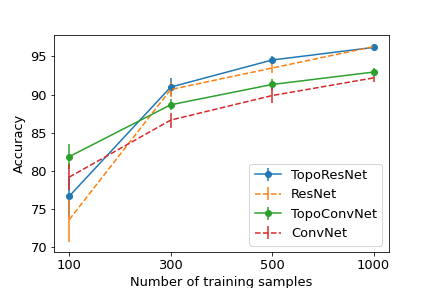}
 &  \hspace{9ex}\includegraphics[width=0.42\linewidth]{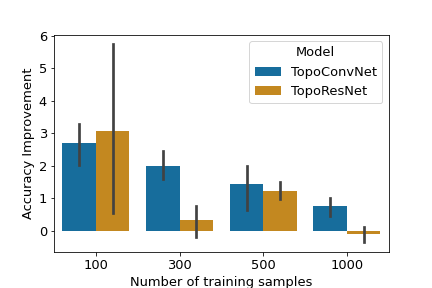}
\\
(a) MNIST&\hspace{9ex}(b) MNIST\\

\includegraphics[width=0.42\linewidth]{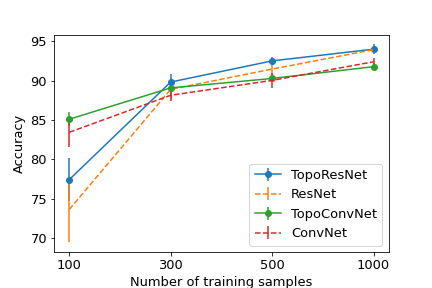}
 &
\hspace{9ex}
 \includegraphics[width=0.42\linewidth]{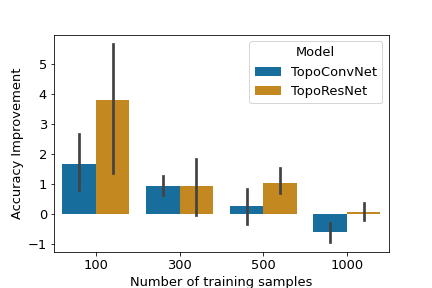}\\
(c) USPS&\hspace{9ex}(d) USPS

\end{tabular}
     \caption{
     A comparison of ConvNet vs TopoConvNet on digits. 
   X axis depicts the number of samples in the training set, Y axis depicts the test accuracy in percentages. 
   }
   \label{topo_conv} 

\end{figure*}

Let X be the input image. The computation of topological features involves three main steps, as follows:
\begin{itemize}

    \vspace{-2ex}
    \item Constructing a cubical complex from the input image: $X \rightarrow K$;
    \vspace{-2ex}
     \item Applying persistent homology by filtering by grayscale values to extract topological features: $K \rightarrow \{\mathcal{D}_0, \mathcal{D}_1\}$;
     \vspace{-2ex}
     \item Performing a vectorization of persistence diagrams: $ \{\mathcal{D}_0, \mathcal{D}_1\} \rightarrow \{V_0^{kq}, V_1^{kq}\}$.
     \vspace{-2ex}
\end{itemize}
As a result, two fixed-length vectors of size $[kq]$ are obtained (one for each feature dimension), with $q$ being the number of binning points and $k$ being the the number of landscapes to be kept. $q$ and $k$ are the hyper parameters of the persistence landscapes that need to be defined in advance. 

\subsection{Landscape Layer}\label{section:land_layer}
As a result of the aforementioned procedure we obtain two finite vectors that summarize the topology of the image. They can be further fed to a Neural Network. In contrast to previous approaches ~\cite{hofer_learning_rep, carriere_perslay_2020, kim_efficient_2020}, that concatenate 0D and 1D persistence landscapes and use the resulting vector as an input to a following layer, we construct separate linear layers for each dimension of the persistent features. This architecture decision is motivated by the difference in the distribution of 0D and 1D features and is particularly beneficial for images with a complex topological structure (e.g., medical patches). 
The architecture of the Landscape Layer is shown in Fig.\ref{fig:landscape_layer}.

The Landscape Layer is fully differentiable and thus can be trained end-to-end using any gradient-based optimizer, e.g., Adam or SGD. However, the topological information alone is usually insufficient to make a competitive classifier. Instead, the topological layer is often used as an addition to the main backbone network.

The influence of the layer is bigger when it is trained jointly with the main network, rather than separately. However, in some setups the influence of adding the layer to the main architecture might be rather harmful. In these cases, the ensemble technique is preferred (see Appendix \ref{section:integration_methods} for more details). 


\section{Experiments}
\label{sec:experiments}


In this section, we extensively evaluate the impact of the topological properties on the classification task for three image datasets:  MNIST~\cite{lecun2010mnist}, USPS~\cite{uspsdataset}, and a medical dataset~\cite{zhang_pathologist-level_2019}. We refer to Appendix \ref{experimental_setup} for more detailed description of the datasets, network structures and training hyperparameters.

The goal of the experiments is to quantitatively analyse the effectiveness of using the topological properties of the dataset. 



We start our experiments by comparing the influence of adding the topological layer with respect to the backbone network complexity (Section \ref{sec:network_complexity}). We first study the effect of adding the Landscape Layer to a simple Convolutional Backbone Network (ConvNet), namely TopoConvNet. We further proceed by taking a more complex backbone network of ResNet, and observe if it can similarly benefit from the topological features. 

The next set of experiments focuses on analysing the impact of using topology with respect to the dataset size (Section \ref{sec:sample_size}). We vary the number of samples available during training, and observe if adding the Landscape Layer to the main network can impact accuracy.

After performing an extensive study on digits datasets with images of a relatively simple topological structure, we further proceed to analyze the influence of the Landscape Layer on a more complex medical dataset (Section \ref{sec:data_complexity}). The complexity of the medical dataset is reflected in a high variability of the topological structures of the patches, which results in significantly bigger influence of the landscape layer. 





%

\begin{table}[t]
\begin{center}

\resizebox{0.48\textwidth}{!}{

\begin{tabular}{|c|c|c|c|c|}
\hline
 Dataset     &ConvNet & TopoConvNet & ResNet & TopoResNet\\
\hline
MNIST &  
86.6$\pm$1.0 & 88.6$\pm$0.7 &  90.6$\pm$0.9 & 91.0$\pm$1.2\\
USPS & 
  88.1$\pm$0.7 & 89.0$\pm$0.5 & 88.9$\pm$1.3& 89.8$\pm$1.1\\
\hline
\end{tabular}}
\vspace{-3ex}
\end{center}
 \caption{Performance comparison with regards to the network complexity, and training size of 300 samples.  The values represent the mean and a std of the classification accuracy over 10 folds.}
    
\vspace{-3.8ex}
\label{table:net_complexity}
\end{table}

\subsection{Network Complexity}\label{sec:network_complexity}

    

To study the effectiveness of incorporating topology with respect to the complexity of the backbone architecture, we show the results of adding a Landscape Layer to a simple baseline of ConvNet. For this set of experiments, we set the training size to the lowest possible number of training samples $n=300$, for which we get reasonably stable results in different runs of the algorithm.

The resulting test accuracy can be found in Table \ref{table:net_complexity}. We report the mean and the standard deviation over 10 runs of the experiment. The results show that for a small dataset, adding the Landscape Layer to ConvNet has a beneficial effect: it improves the accuracy
by 2\% on MNIST and by 1\% on USPS.

Experiments, described above, demonstrated that the topological layer can significantly enhance the performance of a simple neural network.
However, there is no existing analysis on the influence of using topology in a more complex deep neural network, such as ResNet, for a topologically simple dataset. Therefore, we perform a set of experiments to compare the performance of ResNet with its topological counterpart. The results from Table \ref{table:net_complexity} reveal that the impact of the Landscape Layer is marginal on ResNet in comparison to the ConvNet. This confirms that incorporating topological information in the training stage helps improving the recognition accuracy, however, such information can become redundant for more complex backbone networks. 


\subsection{Dataset Size}\label{sec:sample_size}

In this section, we further investigate the influence of incorporating topological information of the dataset in the training stage with respect to different dataset sizes.

We vary the number of samples $n \in \{100, 300, 500, 1000\}$ available during training, and observe that the training size can significantly affect the performance. For small $n$, the performance is sensitive not only to the choice of the samples, but also to the initialization of the network.
Therefore, for a fair comparison between the backbone network and its topological counterpart, we fix the initialisation of their common layers, as well as the data splits. Moreover, to ensure the stability of the reported results, the samples are drawn randomly 10 times for each training size $n$. The test set is always the same throughout these experiments. %

To clarify our findings, in Fig.\ref{topo_conv}(b,d) we further show the performance improvement with respect to training size as a pairwise difference between the accuracies of ConvNet vs. TopoConvNet, and Resnet vs. TopoResnet.

 \begin{figure}[t]
\vspace{-4ex}
\begin{center}
\includegraphics[width=0.9\linewidth]{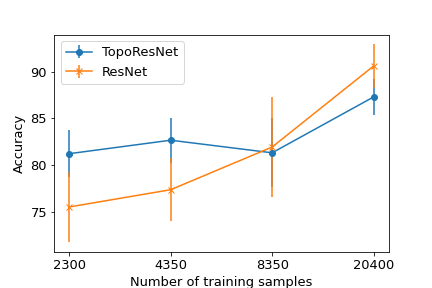}
\end{center}
\vspace{-1ex}
     \caption{
     A comparison of ResNet vs TopoResNet, trained on Medical data.
   X axis depicts the number of samples in the training set, Y axis depicts the test accuracy in percentages. 
   }
   \label{medical_resnet} 
   \vspace{-2ex}
\end{figure}

The resulting plots for MNIST can be found in Fig.\ref{topo_conv}(a,b). The biggest improvement for ConvNet is achieved when $n$ is small: over 2.5\% on average for $n=100$ and around 2\% for $n=300$. With increasing training size, the accuracy improvement of TopoConvNet decreases and falls below 1\% for $n=1,000$. 

Analogous performance trends are observed while training on USPS as shown in Fig.\ref{topo_conv}(c,d): the biggest accuracy improvement of 1.6\% appears while training ConvNet on small subsets of data ($n=100$). Similarly, augmenting the number of training samples leads to a decline in the influence of the Landscape layer. When $n$ is big enough, the Landscape Layer does not help the performance of the main classifier, and in certain cases, e.g., when $n=1,000$, it can even be harmful. 
\begin{table}
\begin{center}
\begin{tabular}{|c|c|c|c|}
\hline
 Dataset     &  ResNet & TopoResNet\\
\hline

MNIST & 97.85$\pm$0.16 &  97.87$\pm$0.33 \\
USPS &  95.78$\pm$0.37& 95.79$\pm$0.38\\
Medical & 75.57 $\pm$3.75 & 81.25 $\pm$2.47\\

\hline
\end{tabular}
\vspace{-2ex}
\end{center}
    \caption{Performance comparison with respect to data complexity, and  training size of $\approx 2K$ samples. The values represent the mean and a std of the classification accuracy over 10 folds.}
    \label{table:data_complexity_comparison}
\vspace{-2ex}
\end{table}
\setlength{\fboxrule}{1pt}

\setlength{\fboxrule}{1pt}
\begin{figure*}
\begin{tabular}{cc}

\hspace{1cm}
\fcolorbox{green}{white}{

\includegraphics[width=0.12\linewidth]{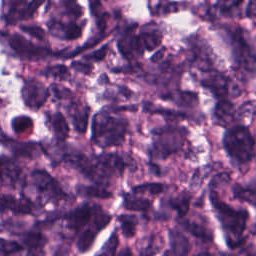}
\includegraphics[width=0.12\linewidth]{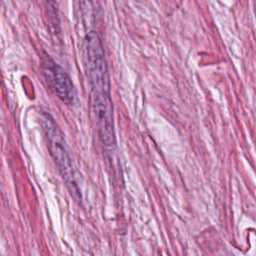}
\includegraphics[width=0.12\linewidth]{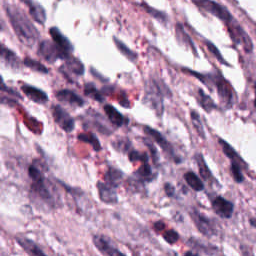}
}
\hspace{1cm}
&
\fcolorbox{green}{white}{
\includegraphics[width=0.12\linewidth]{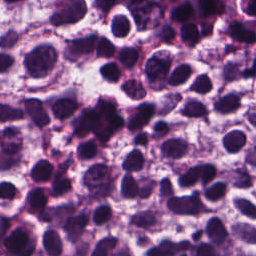}
\includegraphics[width=0.12\linewidth]{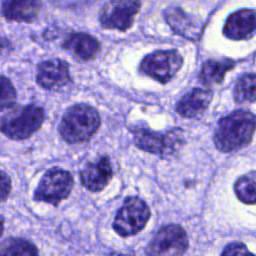}
\includegraphics[width=0.12\linewidth]{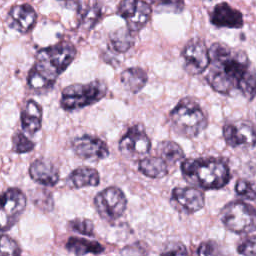}
}

\vspace{0.1cm}

\\(a) Normal &(b) Tumor\\

\fcolorbox{red}{white}{
\includegraphics[width=0.12\linewidth]{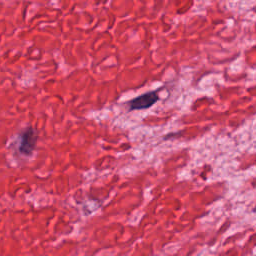}
\includegraphics[width=0.12\linewidth]{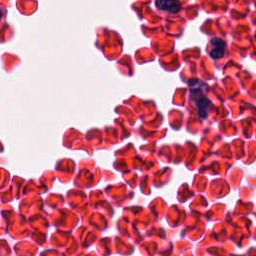}
\includegraphics[width=0.12\linewidth]{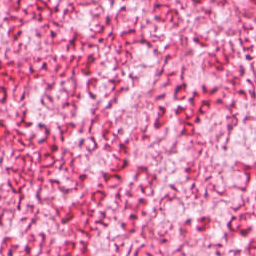}
}
&
\fcolorbox{red}{white}{
\includegraphics[width=0.12\linewidth]{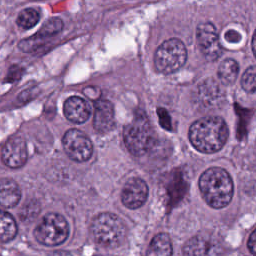}
\includegraphics[width=0.12\linewidth]{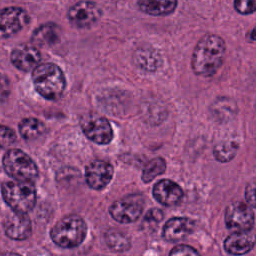}
\includegraphics[width=0.12\linewidth]{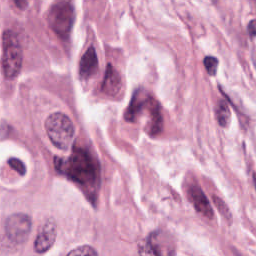}
}

\\(c) Normal&(d) Tumor\\

\end{tabular}
\vspace{-2ex}
     \caption{
     Example of medical patches where adding topology helped (top) and harmed (bottom) the performance. 
     Top: samples correctly classified by TopoResNet and misclassified by ResNet. Bottom: samples correctly classified by ResNet, but misclassified by TopoResNet.   }
   \label{med_patches} 
\end{figure*}

Unsurprisingly, the performance of ResNet and TopoResNet on the smallest $n$ is very poor and even inferior to ConvNet: it quickly overfits on the training data and does not generalise well on the test data. As the training size increases, ResNet starts to significantly outperform ConvNet.

Note that, we are more interested in settings where ResNet is better than ConvNet: $n>=300$, for MNIST and USPS. For both datasets, we observe the trend of having a positive effect of the Landscape Layer when $n$ is small. Moreover, for USPS, when $n$ is big enough, e.g., $n=1,000$, the Landscape Layer does not harm the performance of the ResNet as much as the ConvNet. This may come from the fact that ResNet by itself produces more confident predictions compared to the layer.
Note that, the influence of the Landscape Layer on ResNet does not have the same linear correlation with the training size as observed for ConvNet. For example, the average improvement on MNIST for $n=500$ is bigger than for $n=300$; similarly, the variance of the improvement on USPS for $n=500$ is higher than for $n=300$.

\subsection{Data Complexity}\label{sec:data_complexity}

In the previous two sections, all the experiments have been performed on the digits datasets, that consist of small images with simple topological structure. In this section, 
we report our results on the more topologically diverse medical dataset.

For the medical data, the training size $n$ is defined by the number of Whole Slide Images (WSI) with $n \in \{5, 10, 20, 50\}$. Each WSI has annotated regions of different sizes,  meaning that the actual number of patches in a training set vary depending on the training split. For example, 5 WSIs can indicate from  838 to 3,434 patches in the training set. For more details, please refer to Appendix \ref{experimental_setup}.

Since ConvNet is not deep enough to capture the complex structure of the medical images, we perform our experiments using the Resnet backbone. We compare the performance of ResNet and TopoResNet on the smallest $n=5$ WSIs for the medical dataset. To analyse the effect of incorporating Landscape Layer with respect to data complexity, we further add the results of Resnet and TopoResnet on the same training size of $n=2,300$ samples for MNIST and USPS datasets. The results shown in Table \ref{table:data_complexity_comparison} demonstrate that for a small training size, the medical dataset significantly benefits from the additional information provided by the Landscape Layer. Using TopoResnet, we observe a large performance improvement (5.5\%) for medical data in comparison to no improvement for the topologically simple digits datasets. This finding has a substantial practical application, as some rare diseases have only a few WSIs available for training. 

On medical data, the influence of the Landscape layer, both positive and negative, is significantly larger (Fig.\ref{medical_resnet}). 
The complexity of the medical data is reflected in a high variability of the topological structures of the patches and thus results in more confident predictions.
The biggest improvement is obtained when the training size is small: 5.7\% on average for $n=5$; and 5.3\% on average for $n=10$. 
However, when the training size is big enough, $n=50$, the accuracy of TopoResnet drops on average by 3.3\% compared to Resnet. This result is consistent with our findings for the digits datasets in Section \ref{sec:sample_size}.

\section{Discussion and Future Work}

The improvement from adding topology layer to the network trained on the medical dataset comes primarily from eliminating false positives (Fig.\ref{med_patches}, a). TopoResNet is more robust to color variations; it successfully detects the connectivity of the cells and recognises tumorous cases with distinct and well-separated formations (Fig.\ref{med_patches}, b). Concurrently, TopoResNet tends to confuse some normal patches as being tumorous due to their topological similarity (Fig.\ref{med_patches}, c), and misclassify patches with low color variation (Fig.\ref{med_patches}, d). 
We refer to Appendix \ref{digit_analysis} for a similar analysis of the experiments on the digit datasets.

We observed that when the training size is small, the topological layer trained as a part of the main network can significantly enhance the main classifier. 
However, it might become harmful once enough training samples become available. Such trend is justified by the fact that some of the images belonging to different classes are topologically equivalent. In this case, it might be preferable to
employ a different integration method of the topological layer, which we discuss in more details in Appendix \ref{section:integration_methods}.

Overall, we identify the biggest discovered issue to be the topological inconsistency of the datasets. To tackle this problem, one might design an optimal dataset-specific transformation to extract consistent topological features, which is the focus of our future work.

\textbf{Acknowledgement}\\
This research is supported by the Shenyang Science and Technology Plan Fund (No. 20-201-4-10), the Member Program of Neusoft Research of Intelligent Healthcare Technology, Co. Ltd.(No. NRMP001901)).

\nocite{langley00}

\bibliography{main}

\begin{thebibliography}{24}
\providecommand{\natexlab}[1]{#1}
\providecommand{\url}[1]{\texttt{#1}}
\expandafter\ifx\csname urlstyle\endcsname\relax
  \providecommand{\doi}[1]{doi: #1}\else
  \providecommand{\doi}{doi: \begingroup \urlstyle{rm}\Url}\fi

\bibitem[Adams et~al.(2017)Adams, Emerson, Kirby, Neville, Peterson, Shipman,
  Chepushtanova, Hanson, Motta, and Ziegelmeier]{adamce_persistence_image}
Adams, H., Emerson, T., Kirby, M., Neville, R., Peterson, C., Shipman, P.,
  Chepushtanova, S., Hanson, E., Motta, F., and Ziegelmeier, L.
\newblock Persistence images: A stable vector representation of persistent
  homology.
\newblock \emph{Journal of Machine Learning Research}, 18\penalty0 (8), 2017.
\newblock URL \url{http://jmlr.org/papers/v18/16-337.html}.

\bibitem[Al-Saffar et~al.(2020)Al-Saffar, Bialkowski, Baktashmotlagh, Trakic,
  Guo, and Abbosh]{9274540}
Al-Saffar, A., Bialkowski, A., Baktashmotlagh, M., Trakic, A., Guo, L., and
  Abbosh, A.
\newblock Closing the gap of simulation to reality in electromagnetic imaging
  of brain strokes via deep neural networks.
\newblock \emph{IEEE Transactions on Computational Imaging}, 7:\penalty0
  13--21, 2020.

\bibitem[Bubenik(2015)]{bubenic_landscapes}
Bubenik, P.
\newblock Statistical topological data analysis using persistence landscapes.
\newblock \emph{Journal of Machine Learning Research}, 16\penalty0 (3), 2015.
\newblock URL \url{http://jmlr.org/papers/v16/bubenik15a.html}.

\bibitem[Byrne et~al.(2021)Byrne, Clough, Montana, and
  King]{byrne_persistent_2020}
Byrne, N., Clough, J.~R., Montana, G., and King, A.~P.
\newblock A persistent homology-based topological loss function for multi-class
  cnn segmentation of cardiac mri.
\newblock In \emph{Statistical Atlases and Computational Models of the Heart.
  M{\&}Ms and EMIDEC Challenges}, 2021.
\newblock ISBN 978-3-030-68107-4.

\bibitem[Carriere et~al.(2020)Carriere, Chazal, Ike, Lacombe, Royer, and
  Umeda]{carriere_perslay_2020}
Carriere, M., Chazal, F., Ike, Y., Lacombe, T., Royer, M., and Umeda, Y.
\newblock Perslay: A neural network layer for persistence diagrams and new
  graph topological signatures.
\newblock In \emph{Proceedings of the Twenty Third International Conference on
  Artificial Intelligence and Statistics}, 2020.

\bibitem[Chazal et~al.(2014)Chazal, Fasy, Lecci, Rinaldo, and
  Wasserman]{chazal_persistence_silhouettes}
Chazal, F., Fasy, B.~T., Lecci, F., Rinaldo, A., and Wasserman, L.
\newblock Stochastic convergence of persistence landscapes and silhouettes.
\newblock In \emph{Proceedings of the Thirtieth Annual Symposium on
  Computational Geometry}, 2014.
\newblock ISBN 9781450325943.
\newblock \doi{10.1145/2582112.2582128}.
\newblock URL \url{https://doi.org/10.1145/2582112.2582128}.

\bibitem[Chung et~al.(2018)Chung, Hu, Lawson, and
  Smyth]{chung_topological_2018}
Chung, Y., Hu, C., Lawson, A., and Smyth, C.
\newblock Topological approaches to skin disease image analysis.
\newblock In \emph{IEEE International Conference on Big Data}, 2018.
\newblock \doi{10.1109/BigData.2018.8622175}.
\newblock URL
  \url{https://doi.ieeecomputersociety.org/10.1109/BigData.2018.8622175}.

\bibitem[Clough et~al.(2020)Clough, {\"O}ks{\"u}z, Byrne, Zimmer, Schnabel, and
  King]{clough_topological_2020}
Clough, J., {\"O}ks{\"u}z, I., Byrne, N., Zimmer, V., Schnabel, J., and King,
  A.~P.
\newblock A topological loss function for deep-learning based image
  segmentation using persistent homology.
\newblock \emph{IEEE transactions on pattern analysis and machine
  intelligence}, 2020.

\bibitem[Dey et~al.(2017)Dey, Mandal, and Varcho]{dey_improved_2017}
Dey, T.~K., Mandal, S., and Varcho, W.
\newblock {Improved Image Classification using Topological Persistence}.
\newblock In \emph{Vision, Modeling and Visualization}, 2017.
\newblock ISBN 978-3-03868-049-9.
\newblock \doi{10.2312/vmv.20171272}.

\bibitem[Edelsbrunner \& Harer(2010)Edelsbrunner and Harer]{edelsbrunner_2010}
Edelsbrunner, H. and Harer, J.
\newblock \emph{Computational Topology - an Introduction.}
\newblock American Mathematical Society, 2010.
\newblock ISBN 978-0-8218-4925-5.

\bibitem[Gabrielsson et~al.(2020)Gabrielsson, Nelson, Dwaraknath, and
  Skraba]{bruel-gabrielsson_topology_2019}
Gabrielsson, R.~B., Nelson, B.~J., Dwaraknath, A., and Skraba, P.
\newblock A topology layer for machine learning.
\newblock In \emph{The 23rd International Conference on Artificial Intelligence
  and Statistics, {AISTATS}}, 2020.

\bibitem[He et~al.(2016)He, Zhang, Ren, and Sun]{He2016DeepRL}
He, K., Zhang, X., Ren, S., and Sun, J.
\newblock Deep residual learning for image recognition.
\newblock \emph{2016 IEEE Conference on Computer Vision and Pattern Recognition
  (CVPR)}, pp.\  770--778, 2016.

\bibitem[Hofer et~al.(2017)Hofer, Kwitt, Niethammer, and Uhl]{hofer_deep_2018}
Hofer, C., Kwitt, R., Niethammer, M., and Uhl, A.
\newblock Deep learning with topological signatures.
\newblock In \emph{Advances in Neural Information Processing Systems}, 2017.

\bibitem[Hofer et~al.(2019)Hofer, Kwitt, and Niethammer]{hofer_learning_rep}
Hofer, C.~D., Kwitt, R., and Niethammer, M.
\newblock Learning representations of persistence barcodes.
\newblock \emph{Journal of Machine Learning Research}, 20\penalty0 (126), 2019.
\newblock URL \url{http://jmlr.org/papers/v20/18-358.html}.

\bibitem[Hu et~al.(2019)Hu, Li, Samaras, and Chen]{hu_topology-preserving}
Hu, X., Li, F., Samaras, D., and Chen, C.
\newblock Topology-preserving deep image segmentation.
\newblock In \emph{Advances in Neural Information Processing Systems}, 2019.

\bibitem[{Hull}(1994)]{uspsdataset}
{Hull}, J.~J.
\newblock A database for handwritten text recognition research.
\newblock \emph{IEEE Transactions on Pattern Analysis and Machine
  Intelligence}, 16\penalty0 (5), 1994.
\newblock \doi{10.1109/34.291440}.

\bibitem[Kim et~al.(2020)Kim, Kim, Zaheer, Kim, Chazal, and
  Wasserman]{kim_efficient_2020}
Kim, K., Kim, J., Zaheer, M., Kim, J., Chazal, F., and Wasserman, L.
\newblock Pllay: Efficient topological layer based on persistent landscapes.
\newblock In \emph{Advances in Neural Information Processing Systems}, 2020.

\bibitem[LeCun et~al.(2010)LeCun, Cortes, and Burges]{lecun2010mnist}
LeCun, Y., Cortes, C., and Burges, C.
\newblock Mnist handwritten digit database.
\newblock \emph{ATT Labs [Online]. Available:
  http://yann.lecun.com/exdb/mnist}, 2, 2010.

\bibitem[Moor et~al.(2020)Moor, Horn, Rieck, and
  Borgwardt]{moor_topological_2020}
Moor, M., Horn, M., Rieck, B., and Borgwardt, K.
\newblock Topological autoencoders.
\newblock In \emph{Proceedings of the 37th International Conference on Machine
  Learning}, 2020.
\newblock URL \url{http://proceedings.mlr.press/v119/moor20a.html}.

\bibitem[Qaiser et~al.(2019)Qaiser, Tsang, Taniyama, Sakamoto, Nakane, Epstein,
  and Rajpoot]{qaiser_fast_2018}
Qaiser, T., Tsang, Y., Taniyama, D., Sakamoto, N., Nakane, K., Epstein, D., and
  Rajpoot, N.
\newblock Fast and accurate tumor segmentation of histology images using
  persistent homology and deep convolutional features.
\newblock \emph{Medical Image Analysis}, 55, 2019.
\newblock \doi{10.1016/j.media.2019.03.014}.

\bibitem[{The GUDHI Project}(2021)]{gudhi:urm}
{The GUDHI Project}.
\newblock \emph{{GUDHI} User and Reference Manual}.
\newblock {GUDHI Editorial Board}, {3.4.1} edition, 2021.
\newblock URL \url{https://gudhi.inria.fr/doc/3.4.1/}.

\bibitem[Zhang et~al.(2019)Zhang, Chen, McGough, Xing, Wang, Bui, Xie, Sapkota,
  Cui, Dhillon, Ahmad, Khalil, Dickinson, Shi, Liu, Su, Cai, and
  Yang]{zhang_pathologist-level_2019}
Zhang, Z., Chen, P., McGough, M., Xing, F., Wang, C., Bui, M., Xie, Y.,
  Sapkota, M., Cui, L., Dhillon, J., Ahmad, N., Khalil, F., Dickinson, S., Shi,
  X., Liu, F., Su, H., Cai, J., and Yang, L.
\newblock Pathologist-level interpretable whole-slide cancer diagnosis with
  deep learning.
\newblock \emph{Nature Machine Intelligence}, 1, 2019.
\newblock \doi{10.1038/s42256-019-0052-1}.

\bibitem[Zhao \& Wang(2019)Zhao and Wang]{10.5555/3454287.3455171}
Zhao, Q. and Wang, Y.
\newblock Learning metrics for persistence-based summaries and applications for
  graph classification.
\newblock In \emph{Advances in Neural Information Processing Systems}. Curran
  Associates Inc., 2019.

\bibitem[Zomorodian \& Carlsson(2005)Zomorodian and Carlsson]{zomorodian_2005}
Zomorodian, A. and Carlsson, G.
\newblock Computing persistent homology.
\newblock \emph{Discrete and Computational Geometry}, 33:\penalty0 249--274, 02
  2005.
\newblock \doi{10.1007/s00454-004-1146-y}.

\end{thebibliography}
\bibliographystyle{icml2022}

\newpage
\appendix
\onecolumn

\section{Experimental setup} \label{experimental_setup}

All the topology-related calculations were performed using GUDHI library in Python \cite{gudhi:urm}. 

\subsection{Datasets}
\textbf{MNIST:} The MNIST dataset ~\cite{lecun2010mnist}  has 60,000 training and 10,000 test images, depicting grayscale handwritten digits of relatively small size (28x28 pixels). The images are equally distributed between 10 classes.  The task is to predict a correct label for a given image.

\textbf{USPS:} The USPS dataset ~\cite{uspsdataset} consists of 7,291 train and 2,007 test images of grayscale handwritten digits, automatically scanned from envelopes by the U.S. Postal Service. Each image is 16x16 pixels. The task is to classify a given image into one of 10 classes.

\textbf{Medical dataset:} The Medical dataset, collected by Zhang et al.\cite{zhang_pathologist-level_2019} contains haematoxylin and eosin (H\&E) stained whole slide images (WSI) with bladder cancer. Each WSI is accompanied with a pixel-wise annotation of tumorous and non-tumorous regions. The average size of one WSI is 80,386x59,143 pixels. In the preprocessing stage, patches 
of 256x256 pixels are mined from each WSI. Since each WSI has different number and different size of annotated regions, we randomly sample at most 700 patches per WSI. The class distribution in both training and test sets is balanced. The task is to classify the patches into tumorous and normal.

For experiments, described in Section \ref{sec:sample_size} and ~\cite{9274540},  we restrict the number of samples $n$, available during training: $n \in \{100, 300, 500, 1000\}$. The choice of the samples may significantly affect the performance. For this reason the samples are drawn randomly 10 times for each $n$. The test set is always the same throughout all the experiments. 

\subsection{Networks}
   
 The ConvNet consists of one convolutional layer, followed by two fully connected layers. We apply ReLU activation function after the first two layers, 2D maxpooling after the convolutional layer and a dropout with p=0.2 after the first fully connected layer. 
As a large network we chose Resnet-26, described in \cite{He2016DeepRL}. 

The performance of the baselines is compared to their topological counterparts, that have an additional Landscape Layer with $k=3$, $q=50$ for MNIST, $k=2$, $q=50$ for USPS and $k=30$, $q=100$ for a medical dataset. 

In all the setups, networks are trained using Adam optimizer with learning rate $lr=0.001$ and  batch size $b\_size = 32$.

\begin{figure}[h]
\begin{center}
\includegraphics[width=0.5\linewidth]{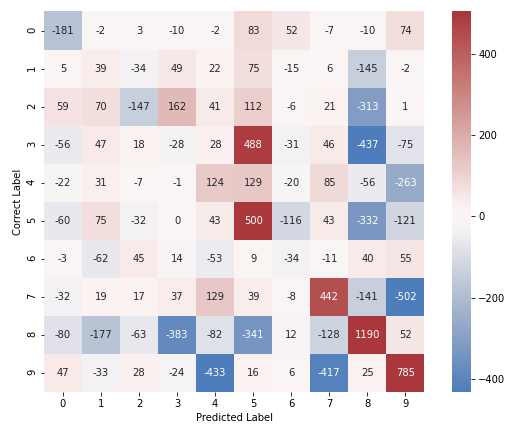}
\end{center}
 \caption{
 Confusion matrix improvement for MNIST with 100 training samples, calculated as conf$_t$ - conf$_c$, with conf$_t$ being the confusion matrix of ConvNet, and conf$_c$ being the the confusion matrix of TopoConvNet. 
 }
   \label{confusion_matrix} 
\end{figure}

\section{Digit experiment analysis} \label{digit_analysis}

\begin{table}
\begin{center}
\begin{tabular}{|c|c|c|c|c|c|c|c|c|c|c|}
\hline
 Dataset     &   \multicolumn{10}{c|}{Classes}\\
\hline
     &0& 1 & 2 &3 &4&5&6&7&8&9 \\
\hline
MNIST
& -1.97
& 0.32
& -1.87
& -0.45
& 1.72
& 7.98
& -0.37
& 5.29
& 16.93
& 11.24
\\
\hline
USPS
& 0.4
& 0.99
& 0.93
& 0.79
& 10.66
& -0.86
& -1.53
& 4.72
& 5.02
& 1.57
\\
\hline
\end{tabular}
\end{center}
    \caption{Percentage gain for training with LandscapeLayer for $n=100$}
    \label{fig:percentage_gain}
\end{table}

MNIST and USPS are composed of handwritten digits, distributed among 10 classes.
Due to the simplicity of these datasets, we can infer the ground truth of the topology of each class, valid for the majority of the images. For example, all digits have only one persistent connected component; digits 1,3,4,5 and 7 have zero holes, digits 0,6,9 have one hole, and digit 8 has two holes. 
Note that this ground truth is not available during training; it is only used to analyse the results.


The accuracy improvement described in Section \ref{sec:experiments} comes from the Landscape Layer. 
In order to validate the origins of this improvement, we analyze the changes in per-class predictions. Fig.\ref{confusion_matrix} depicts the difference between the confusion matrix of TopoConvNet and the one of ConvNet. Intuitively, we would like to have more positive values in the diagonal, representing that TopoConvNet increases the number of correct predictions for each class, and see more negative values off the diagonal, which implies that TopoConvNet has fewer misclassifications than ConvNet.

As demonstrated in Fig.\ref{confusion_matrix}, the biggest enhancement is achieved for the digit 8, as it is the only class with the unique topological signature (one connected component and two holes). The performance for digits 9, 7 and 5 has also increased. For example, TopoConvNet better recognises digit 9 from digits 4 and 7 as digit 9 has one hole, while digits 4 and 7 have no holes.
On the other hand, digits 0 and 2 have experienced a slight deterioration in their classification. 
These experimental results are consistent with the ground truth, described in the beginning of this section.

Finally, a per-class percentage gain shown in Table \ref{fig:percentage_gain} demonstrates that for small training size of $n=100$, adding the Landscape Layer can significantly increase the performance for some classes, while slightly harming the others.

\section{Integration methods}\label{section:integration_methods}

In this section, we analyze the effect of incorporating topological information in the training stage with respect to the integration method.

\begin{figure*}
\begin{tabular}{c|c}
\includegraphics[width=0.46\linewidth]{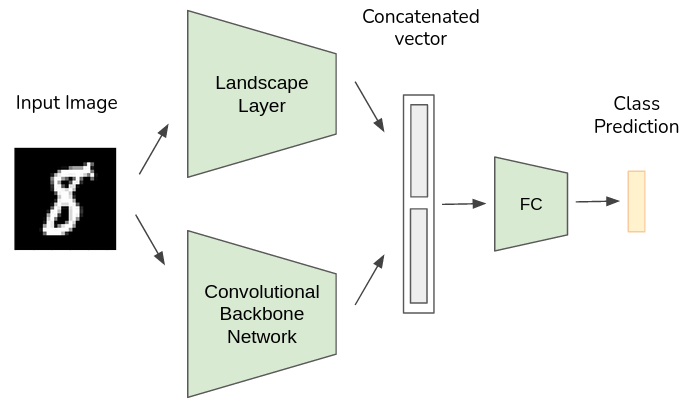}
 & \includegraphics[width=0.46\linewidth]{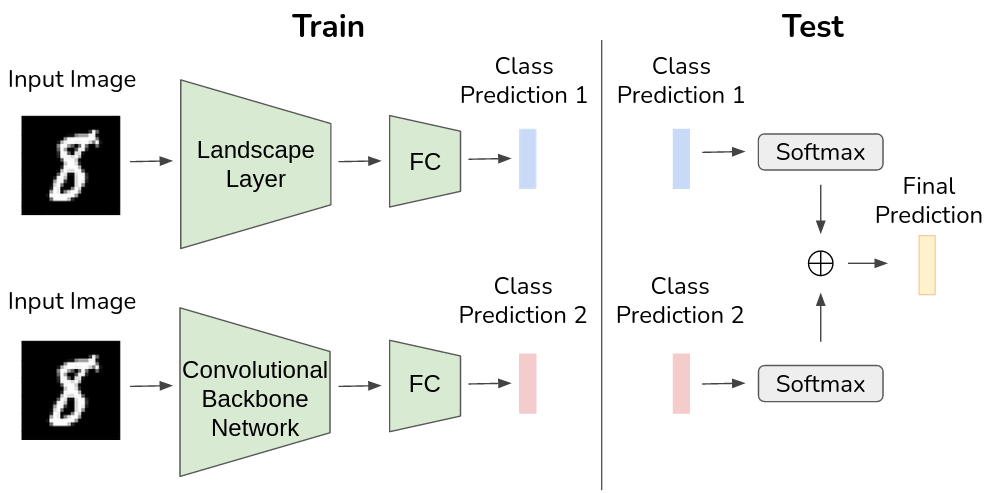}
\\
(a) Topological Neural Network architecture. & (b) Ensemble strategy. \\ 

\end{tabular}
     \caption{
    (a)  The Landscape Layer and the Convolutional Network are combined into one architecture and trained jointly by backpropogating through both branches, as the Landscape Layer is fully differentiable. (b) The Landscape Network and the Convolutional Network are trained independently. In inference time, the predictions from two networks are combined to form a final prediction.
   }
   \label{fig:topoNN_arch} 

\end{figure*}

One way to integrate the layer into a ML pipeline is shown in Fig.\ref{fig:topoNN_arch}(a). Here, the input batch is being passed through the Landscape Layer and a Convolutional Backbone Network, cut after the last convolutional layer. The output of the two is concatenated and is being further passed to a fully connected layer (FC), that in turn returns a final prediction. The described topological network, called TopoNet, is trained with a cross-entropy loss, that backpropogates through both branches.

The other way to use the layer within a ML pipeline is to train the backbone network and the landscape classifier separately, as shown in Fig.\ref{fig:topoNN_arch}(b). At inference time, the predictions from both networks $p_1, p_2$ are combined into a final prediction $p$ as follows: $p$ = argmax(softmax($p_1$) + softmax($p_2$)). We call this the ensemble strategy.

The influence of the layer is bigger when it is trained jointly with the main network, rather than separately. However, in some setups the influence of adding the layer to the main architecture might be rather harmful. In these cases, the ensemble technique is preferred, as TopoNet only affects the predictions it is certain about.

We select the setups where TopoNet showed inferior results to Net (i.e., the network without topological layer). For these setups, we construct the Landscape network, that consists of the Landscape Layer, followed by a fully connected layer. The Landscape Network is further trained using SGD with adaptive learning rate starting at $lr=0.01$. In inference time, the predictions from the backbone network are combined with predictions from the Landscape network.

The performance of the resulting Ensemble method is compared with the TopoNet in Table \ref{table:integration_comparison}. The results demonstrate that unlike TopoNet, the Ensemble never harms the performance of the main classifier, and achieves slight improvements of around 0.7\% for the medical dataset.

\begin{table*}
\begin{center}
\begin{tabular}{|c|c|c|c|c|c|c|}
\hline
 Dataset     & Backbone type &Net & TopoNet& Lanscape Net & Ensemble \\
\hline
USPS &ConvNet& 92.4$\pm$0.49& 91.78$\pm$0.45 & 49.73$\pm$0.77  & 92.7$\pm$0.58 \\
Medical &ResNet& 90.59 $\pm$2.3 & 87.29$\pm$1.94 & 83.26$\pm$1.65 & 91.28$\pm$1.6\\
\hline
\end{tabular}

\end{center}
    \caption{Performance comparison with regards to the integration method. The values represent the mean and a std of the classification accuracy over 10 folds.}
    \label{table:integration_comparison}
\end{table*}





\end{document}